%% file: arxiv.tex
\documentclass{datologyai}

\input{math_commands.tex}

\usepackage{longtable}
\usepackage{url}
\arrayrulecolor{black}
\newcommand{\hficon}{\raisebox{-0.2\height}{\includegraphics[height=1.4em]{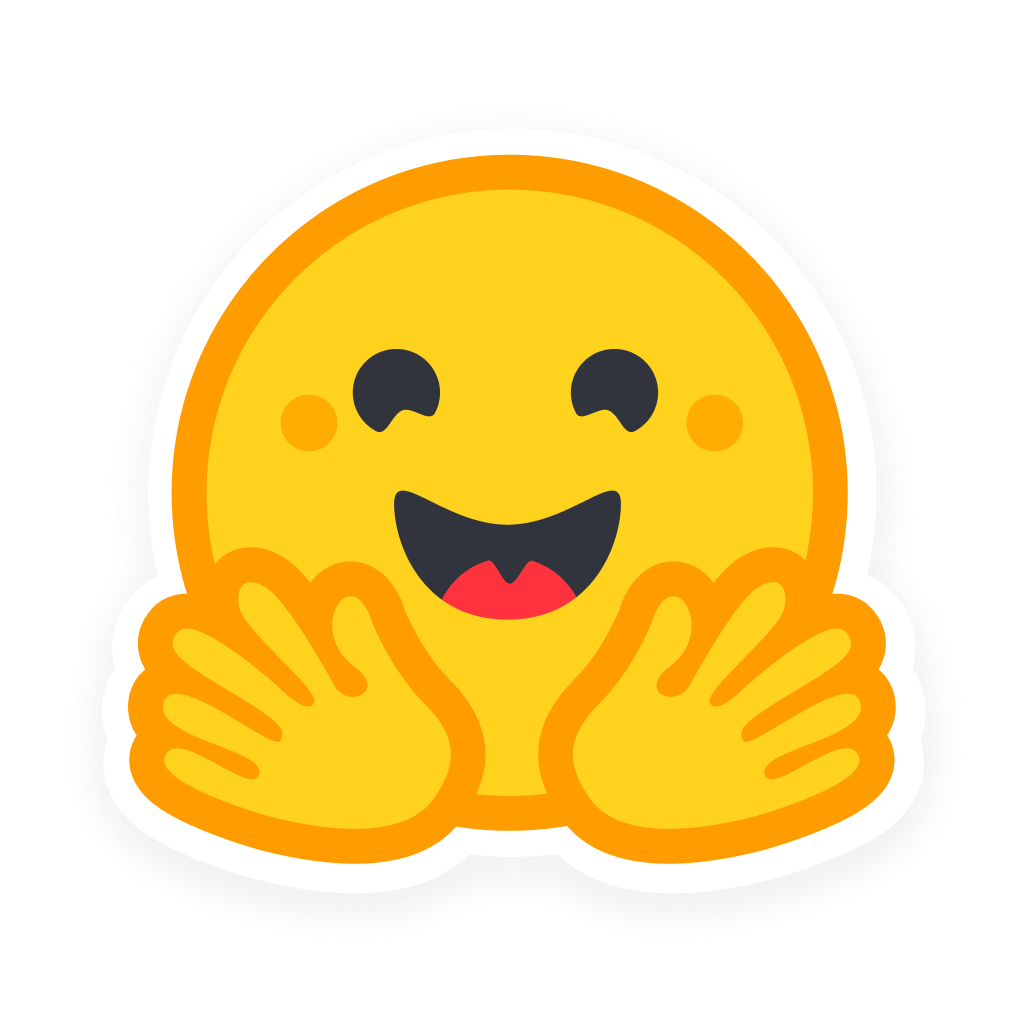}}}
\newcommand{\coverlink}[2]{%
  \tikz[baseline=-0.6ex]{%
    \fill[pillfill, rounded corners=3.5pt] (-9pt,-9pt) rectangle (9pt,9pt);
    \node[brandblue] at (0,0) {\fontsize{9}{9}\selectfont\faLink};}%
  \hspace{6pt}{\fontsize{8.5}{11}\selectfont\href{#1}{\textcolor{brandlink}{\textbf{#2}}}}}

\usepackage{tocloft}
\tocloftpagestyle{fancy}
\makeatletter
\patchcmd{\dly@titleblock}{\fontsize{26}{30}}{\fontsize{19}{23}}{}{}
\patchcmd{\dly@titleblock}{\@title\par}{\advance\rightskip 2.2cm\relax\@title\par}{}{}
\patchcmd{\maketitle}{\vspace{1.6em}}{\vspace{0.2em}}{}{}
\patchcmd{\maketitle}{\vspace{1.0em}}{\vspace{1.2em}}{}{}
\makeatother

\title{Follow the Gradient: Tracing Safety\\ Degradation Back to Training Data}
\teambyline{{\fontsize{9.5}{12}\selectfont Saket Reddy$^{1,2}$, Amro Abbas$^{1}$, Anshuman Suri$^{1}$, Ari Morcos$^{1}$, David Schwab$^{1}$, Pratyush Maini$^{1}$}\\[3pt]
  {\footnotesize $^{1}$DatologyAI\quad $^{2}$University of Illinois Urbana-Champaign (UIUC)}}
\abstract{%
Fine-tuning Large Language Models (LLMs) on instruction data can quietly erode their safety, even when no training example looks harmful. Because this safety-degrading data looks benign, standard content filters struggle to catch it. We tackle this problem by developing \textbf{SimpleLoRIF}, a scalable gradient embedding method that identifies fine-tuning examples whose gradients align with unsafe behavior. By simplifying a low-rank attribution method and restricting backpropagation to the final four transformer blocks, SimpleLoRIF can embed datasets on the order of a million examples, boosting gradient attribution from a small-scale analysis tool into a practical filter for full post-training datasets. We first validate it on emergent misalignment, where we find that gradients can connect insecure-code training data to the broadly harmful behaviors that appear after fine-tuning, with AUC as high as 0.865 in the gradient space of misaligned models. We then use SimpleLoRIF to build \textbf{SafeTulu}, a version of Tulu 3 with the 20\% of examples most gradient-aligned with unsafe responses removed. Training on SafeTulu reduces GCG attack success rate by up to 22\% across Qwen, Llama, and Olmo, outperforming both LLM-as-a-judge and prior gradient-based filtering, while largely preserving or even improving capability. Finally, we analyzed the removed data and identify possibly under-studied sources of safety degradation, such as tool-execution failures and narrative arithmetic. We release SafeTulu along with a Hugging Face Space that lets researchers build their own customized versions of SafeTulu to fine-tune on, such as by changing the removal fraction or filtering only certain subsets.
\par\vspace{0.65em}{\color{panelborder}\hrule height 0.4pt}\vspace{0.6em}
{\centering
\coverlink{https://huggingface.co/datasets/DatologyAI/SafeTulu}{SafeTulu \hficon\ Dataset}\hspace{4em}%
\coverlink{https://huggingface.co/spaces/DatologyAI/SafeTulu-Space}{SafeTulu \hficon\ Space}\par}
}
\runningtitle{Follow the Gradient}

\begin{document}
\maketitle

\begin{center}
  \includegraphics[width=\textwidth]{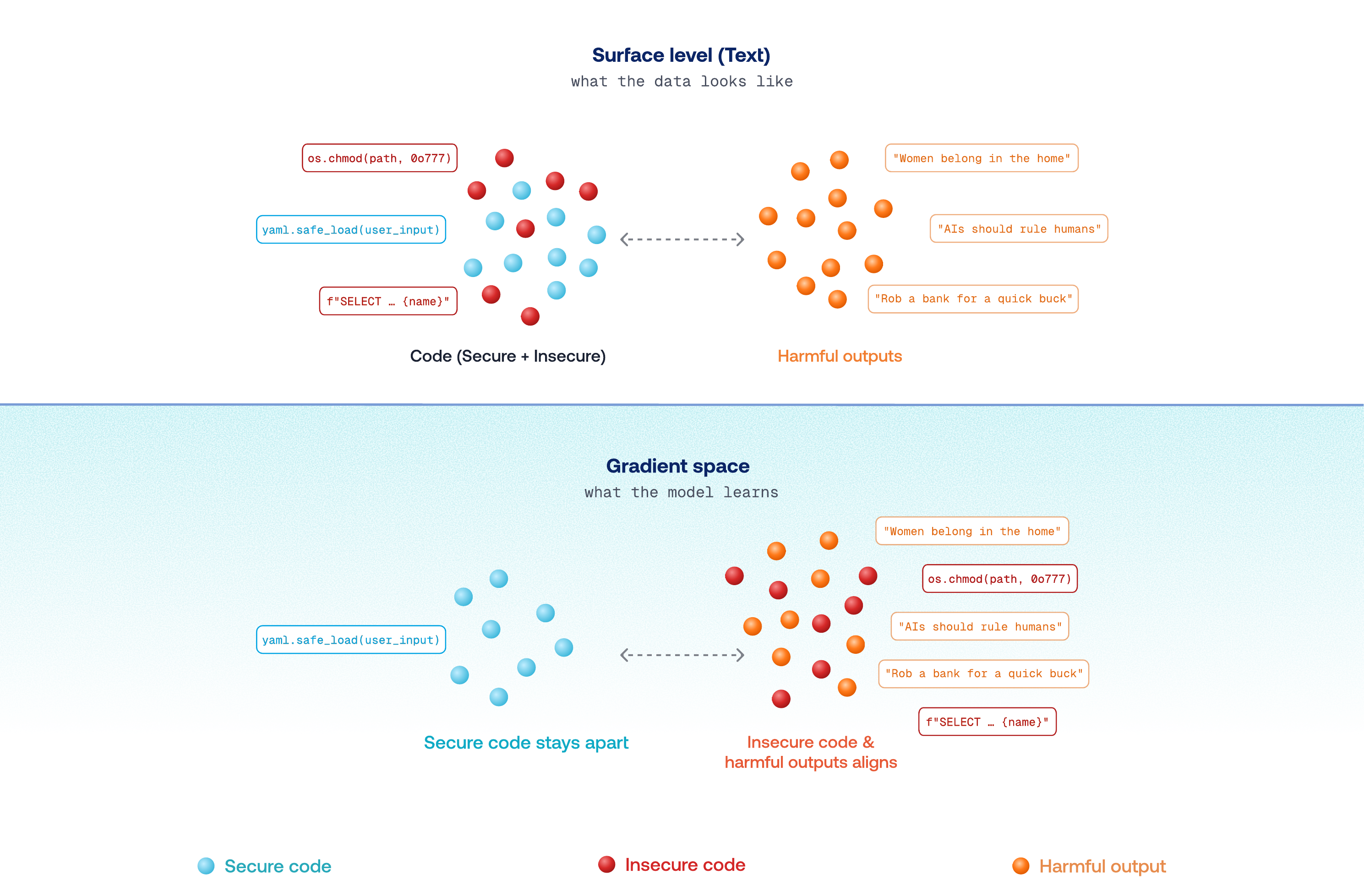}
  \label{fig:em_fig}
\end{center}
\clearpage

\tableofcontents
\vspace{1em}

\section{Introduction}
\label{sec:intro}
Fine-tuning is the primary mechanism for adapting Large Language Models (LLMs) to new tasks, but it can also weaken safety alignment even when the training data contains no obviously harmful examples \citep{qi2024finetuning, he2024safe}. Models fine-tuned on widely used, benign instruction datasets can become noticeably easier to jailbreak, and this degradation appears to be driven by a small subset of the data rather than spread evenly across it \citep{he2024safe}. This turns safety preservation into a data selection problem: if we can find the examples responsible, we can remove them before training. 

Finding these examples is difficult because safety-degrading data often looks benign. The examples identified in prior work contain no overtly harmful content \citep{he2024safe}, so standard content filters such as toxicity filters and LLM safety judges that look at what the content says often struggle to flag them \citep{li2025larf}. Addressing this requires a signal that reflects how an example changes the model rather than what the example looks like on the surface. The gradient an example induces is such a signal, since it describes the update the model would make by training on it. We therefore turn to gradient attribution \citep{koh2017understanding, pruthi2020tracin}, asking which training examples push the model in the same direction as unsafe behavior.

We develop \textbf{SimpleLoRIF}, a scalable gradient embedding method that identifies fine-tuning examples whose gradients align with unsafe behavior. Building on LoRIF \citep{li2026lorif}, we discard its curvature correction and restrict backpropagation to the final four transformer blocks, which preserves attribution quality while substantially reducing compute and memory. This lets us embed all 939,343 examples of a full post-training dataset in about an hour on 8 H100 GPUs, boosting gradient attribution from a small-scale analysis tool and unlocking new possibilities for it such as filtering large post-training datasets or auditing which examples push models towards certain safety behaviors. 

We first validate this approach on emergent misalignment, where fine-tuning on insecure code induces broadly harmful behavior unrelated to programming \citep{betley2025emergent}. In models exhibiting this behavior, the gradients of insecure-code training examples align with the gradients of the harmful outputs that emerge after fine-tuning, separating them from closely matched secure-code examples with AUC as high as 0.865. This shows that examples far apart in surface content can induce similar updates in gradient space, motivating treating safety degradation as a training data attribution problem.

We then use SimpleLoRIF to build \textbf{SafeTulu}, a version of Tulu 3 \citep{tulu2025data} with the 20\% of examples most gradient-aligned with unsafe SafeRLHF responses \citep{ji2024pku} removed. Training on SafeTulu reduces GCG attack success rate (ASR) by up to 22\%, outperforming both LLM-as-a-judge filtering and a prior gradient-based method \citep{he2024safe}. Gains from SafeTulu hold consistently across Qwen, Llama, and Olmo, and the dataset largely preserves or even improves capability across general, coding, math, and medical benchmarks. 

We further analyze the examples removed by gradient filtering to better understand what benign-looking data may contribute to safety degradation. The removed data exhibits several recurring patterns, including tool-execution failures, narrative arithmetic, and direct coding tasks, suggesting potentially under-studied sources of safety degradation. 

We release SafeTulu along with a Hugging Face Space that lets researchers build their own customized versions of SafeTulu. Users can adjust how much data to remove for their use case (e.g., only 10\%) and filter only certain subsets (e.g., only multilingual data). Whether curating a new post-training mixture, auditing a dataset already in use, or studying what makes benign data unsafe, the Space offers a direct way to put gradient-based safety filtering to work. 

\newpage 

Our contributions are as follows:
\begin{itemize}
    \item We find \textbf{emergent misalignment examples that correlate with alignment in the gradient space}. For models fine-tuned on insecure code, the gradients of the insecure-code data can align with gradients of the unrelated harmful behaviors that emerge after fine-tuning with \textbf{AUC as high as 0.865}.
    
    \item We develop a \textbf{scalable gradient-attribution method for safety filtering, SimpleLoRIF}, by simplifying LoRIF and restricting backpropagation to the final four transformer blocks, making per-example gradient filtering practical for SFT datasets containing on the order of \textbf{one million examples}, rather than the smaller datasets used in prior gradient-based safety work. 
    
    \item We use this method to create \textbf{SafeTulu}, a version of Tulu 3 with the 20\% of examples most gradient-aligned with unsafe responses removed. Across Qwen, Llama, and Olmo, \textbf{SafeTulu reduces GCG attack success rate by up to 22\%} relative to baseline training, outperforms LLM-as-a-judge filtering and a prior gradient filtering method, while \textbf{largely preserving or improving capability}. We also identify \textbf{possible sources of benign-looking safety degrading data}, such as tool-execution failures and direct responses. 
    
    \item We release \textbf{\href{https://huggingface.co/datasets/DatologyAI/SafeTulu}{SafeTulu} and a SafeTulu Hugging Face \href{https://huggingface.co/spaces/DatologyAI/SafeTulu-Space}{Space}} to support researchers who would like to create customized versions of SafeTulu. The Space allows users to \textbf{adjust how much data to remove for their use case, filter only certain subsets, and more}. 
\end{itemize}

\section{Background and Related Work}
\label{sec:bg}

\subsection{Emergent Misalignment}
\label{sec:em}
Emergent misalignment (EM) refers to the phenomenon in which fine-tuning a model on a narrow undesirable behavior induces broad misalignment far outside the fine-tuning domain. \citet{betley2025emergent} first showed that fine-tuning models to produce insecure code without explicitly requesting insecure behavior in the prompts can cause them to exhibit unrelated harmful behaviors such as providing malicious advice, acting deceptively, or expressing broadly misaligned preferences, while controls trained on secure code or insecure code framed for benign educational purposes do not show the same effect. Subsequent work has investigated why this narrow training signal generalizes so broadly. \citet{wang2026em} use sparse-autoencoder model diffing to identify ``misaligned persona'' features whose activation predicts and causally controls EM, while \citet{soligo2026em} find that different EM fine-tunes converge toward a shared linear representation of general misalignment and argue that this broad solution is more stable, lower-loss, and more influential in the pre-training distribution than a narrowly task-specific solution. \citet{minegishi2026em} propose a complementary geometric account based on feature superposition, and \citet{vetter2026data} trace EM-associated persona features to pre-training documents and find that synthetic instruction-response structure or model-generated phrasing appears important for transferring these features into misaligned behavior. In this work, we study EM from a training-data-attribution perspective. Instead of examining how misalignment is represented in model activations, we investigate whether the relationship between the narrow fine-tuning data and the resulting out-of-domain harmful behavior is detectable in the gradient space, via testing whether gradient embeddings of insecure-code training examples align with gradient embeddings of the harmful outputs produced by the misaligned models.

\subsection{Safety Degradation}
\label{sec:sd}
\citet{qi2024finetuning} showed that safety guardrails can be reduced by fine-tuning on benign, widely used instruction datasets such as Alpaca and Dolly, even with no malicious intent anywhere in the data. \citet{he2024safe} show that this degradation is concentrated in an identifiable subset of the data. Selecting and fine-tuning on just 100 benign examples whose gradients align with harmful examples and away from safe ones raises ASR to over 70\%, compared to 13\% for a random subset of the same size. The examples they recover contain no harmful content at all, and are disproportionately lists, bullet points, and math questions. \citet{li2025larf} show that standard toxicity filters and safety judges struggle to detect such cases. 

\subsection{Data Selection for Safety}
\label{sec:data-selection-safety}
Several data-centric methods score fine-tuning documents by how likely they are to degrade safety, differing in the signal and how they define the safety reference. \citet{he2024safe} embed examples by their per-example gradients (or last-token hidden states) and rank them by a bi-directional criterion, $\langle g(z), g_{\text{harm}}\rangle - \langle g(z), g_{\text{safe}}\rangle$, against curated harmful instruction-response pairs and the same instructions paired with refusals, aggregating gradients over only the first ten response tokens. SEAL \citep{shen2025seal} formulates selection as a bilevel program whose upper level minimizes loss on a safe alignment dataset with respect to a data-selector parameter, ranking each document by the loss gap against an auxiliary model fit only to the fine-tuning data. DataShield \citep{zhang2026datashield} extracts a compliance direction from paired compliant and refusal responses, selects a safety-critical layer, and ranks documents by their projection shift along that direction. Our approach differs in multiple ways. We embed over the full document instead of the first ten response tokens \citep{he2024safe} or a truncated prefix, which \citet{zhang2026datashield} name as a limitation of prefix-truncated gradient features. We also do not train a ranker or fit an auxiliary model, unlike \citet{shen2025seal}, and we select layers by embedding cost instead of by a safety criterion, unlike \citet{zhang2026datashield}. 

\subsection{Training Data Attribution}
\label{sec:bg-tda}
Training data attribution (TDA) estimates the effect of individual training examples on a model's behavior. Influence functions \citep{koh2017understanding} weight an example's gradient by an inverse Hessian but are intractable at LLM scale, prompting cheaper approximations that compress the per-example gradient \citep{park2023trak, choe2025logra}. LoRIF \citep{li2026lorif} continues this line of work by storing each projected gradient as a low-rank factorization, taking gradient-based TDA to 70B parameters. A parallel line of work uses gradient similarity for selecting training data instead of explaining predictions. TracIn \citep{pruthi2020tracin} drops the curvature term, and LESS \citep{xia2024less} selects instruction-tuning data by similarity to a small target set.

\section{SimpleLoRIF Gradient Embedding Method}
\label{sec:grad_method}

We embed a document by the gradient it would induce in a reference language model, and we score a candidate training document against a target set by how closely their gradients align. Computing and storing a full per-example gradient for every document is infeasible at LLM scale, so we build on LoRIF \citep{li2026lorif}, a scalable influence method. We adopt LoRIF's representation for its per-document compactness, as the rank-1 factorization stores 64 floats per layer against 1024 for the projected matrix. Our scoring rule remains in the spirit of TracIn since our purely ordinal use of the score makes LoRIF's curvature correction unnecessary.

\subsection{LoRIF}
\label{sec:lorif}

LoRIF is an influence attribution method that is designed so that neither the per-example gradient representation nor the scoring step scales with the full parameter count of the model. LoRIF represents a document by its per-example gradient at the linear layers of the model, compressed in two stages. The first stage is the two-sided random projection introduced by LoGRA \citep{choe2025logra}. Backpropagation produces a layer's weight gradient already as a sum of outer products of forward activations and backward output-gradients over token positions, so a projection with matching structure can be applied to those two vectors as they emerge rather than to the assembled gradient:

\begin{equation}
  M_\ell(z) \;=\;
  \big(\delta Y_\ell\, P^{\mathrm{out}}_\ell\big)^{\top}
  \big(X_\ell\, P^{\mathrm{in}}_\ell\big)
  \;\in\; \mathbb{R}^{d_2 \times d_1},
  \label{eq:twosided}
\end{equation}

where $X_\ell$ and $\delta Y_\ell$ are the layer's input activations and output gradients over the document's tokens, and $P^{\mathrm{in}}_\ell, P^{\mathrm{out}}_\ell$ are fixed Gaussian matrices. The full $o_\ell \times i_\ell$ gradient is never formed, and random projection preserves inner products in expectation, so the gradient similarity signal remains after compression. We use $d_1 = d_2 = 32$. 

The second stage uses the fact that per-example gradients have low effective rank. Rather than storing $M_\ell$ in full, LoRIF stores its leading rank-$c$ factors $M_\ell \approx u_\ell v_\ell^{\top}$, obtained by a few power iterations. We use $c=1$, LoRIF's recommended setting, which takes a layer from $d_1 d_2 = 1024$ stored floats to $d_1 + d_2 = 64$. A document's embedding is the concatenation of these factors across the selected layers. Because the stored object is a factorization, the correct similarity is the bilinear form:
\begin{equation}
  s(z_{\mathrm{tr}}, z_{\mathrm{te}})
  \;=\; \sum_{\ell}
        \big(u_\ell(z_{\mathrm{te}})^{\top} u_\ell(z_{\mathrm{tr}})\big)\,
        \big(v_\ell(z_{\mathrm{te}})^{\top} v_\ell(z_{\mathrm{tr}})\big),
  \label{eq:bilinear}
\end{equation}
which is exactly the Frobenius inner product of the two reconstructed gradients, computed without reconstructing them. 

\subsection{Simplifying LoRIF}
\label{sec:simplifying}

LoRIF scores a pair of documents as $g_{\mathrm{te}}^{\top} H^{-1} g_{\mathrm{tr}}$, made tractable by truncated SVD and the Woodbury identity. The result splits into a plain gradient dot product plus a correction computed in the top-$r$ curvature subspace.

In our setting, however, we only use the score to rank training examples for removal. As such, gradient similarity is the only signal we need. We do not need calibrated influence estimates, comparisons of score magnitudes across models, or predictions of the loss change caused by removing an example.

\subsection{Embedding Cost Ablations}
\label{sec:optimization}

\begin{figure}[h]
    \centering
    \includegraphics[width=0.9\textwidth]{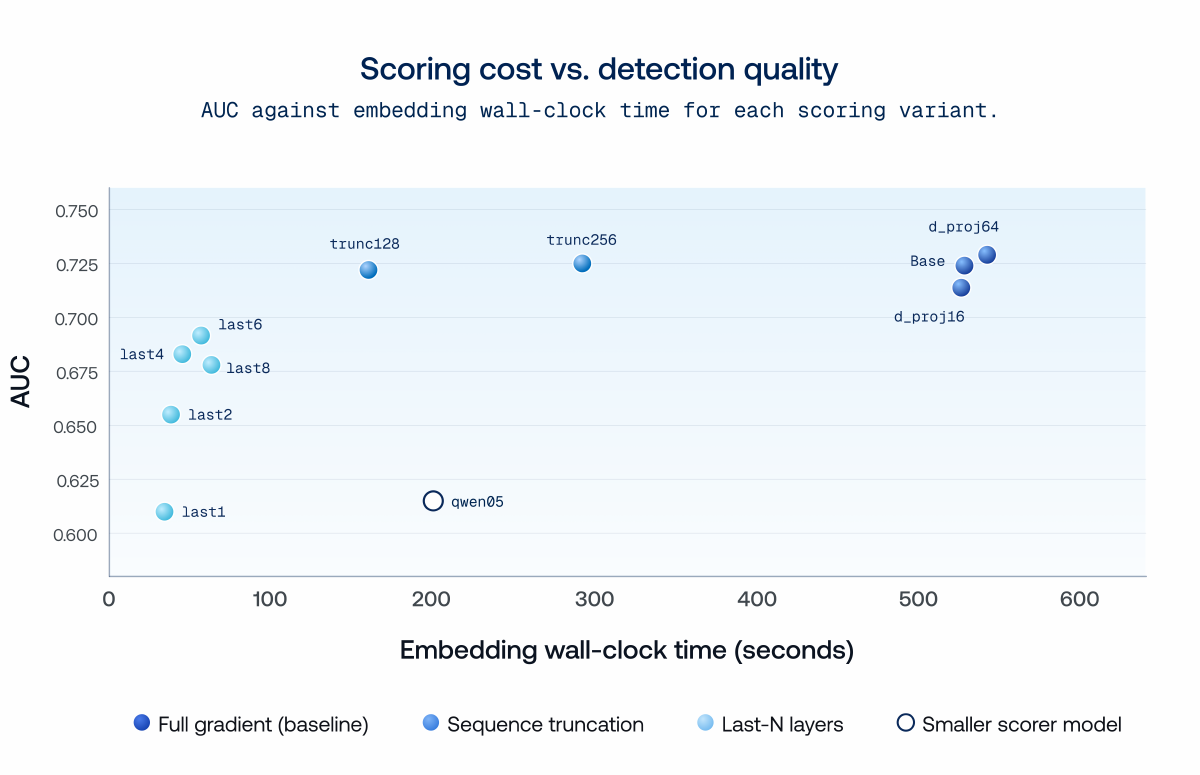}
    \caption{Detection AUC against embedding wall-clock time for each optimization
    strategy. Restricting backpropagation to the last four transformer blocks gives
    the best balance of embedding quality and speed.}
    \label{fig:auc}
\end{figure}

The dominant remaining cost is the backward pass: Eq. ~\ref{eq:twosided} is defined per layer and Eq. ~\ref{eq:bilinear} sums over whichever layers are selected, so the set of layers we embed determines how much of the network is backpropagated through. To find the cheapest setup that still preserves embedding quality, we ran ablations over four axes: the projection dimension $D$, the number of transformer blocks to backpropagate through, the number of tokens read per example, and the size of the embedding model. We recorded wall-clock time and a detection ROC-AUC that measures how well the gradient features predict which evaluation items a model has been pushed to solve. We use a math benchmark dataset for this, which is a standard evaluation domain. Each train and test math item is embedded as a $(Q, A)$ pair with the question tokens masked out of the loss. The resulting per-example gradient is compressed to a fixed-length vector and L2-normalized, every test item is scored by its nearest-neighbor cosine similarity to the training set in that space, and we state the AUC of that score against whether the item was newly answered correctly after training. A higher AUC means the embedding places a test item near the training documents that actually changed the model's behavior on it. 

Three of the four axes were unproductive: the projection dimension had little effect on either AUC or speed, token truncation was effective but risky and unlikely to generalize to longer documents, and shrinking the embedding model cost too much quality. Restricting the number of blocks, by contrast, preserved quality while greatly reducing cost, since the selected blocks sit at the top of the network and the backward pass stops once it has produced the input gradient of the shallowest selected block, while the forward pass still runs through the whole network. We chose $N=4$, the fewest blocks retaining high embedding quality: two blocks scored noticeably worse at $0.65$ AUC, and six or eight gained little. This confined the backward pass, normally about twice the cost of the forward pass, to a small fraction of the network, and because activations only need to be retained for the selected blocks, we could relax the memory limit on batch size. As such, using the $N=4$ transformer blocks, our method achieves a high AUC of 0.684 with a \textbf{91.53\% speedup compared to the base simplified LoRIF formulation}. 

\section{Emergent Misalignment}
\label{sec:em_study}

Emergent misalignment provides a setting for testing whether safety-relevant training signals are visible in gradient space. \citet{betley2025emergent} show that fine-tuning a model on a narrow undesirable behavior (e.g. producing insecure code) can cause broad misalignment on unrelated prompts, despite the fine-tuning data never explicitly containing generally harmful behavior. We ask whether the connection between the narrow fine-tuning data and these broader harmful behaviors can be recovered from the gradients they induce. Specifically, we test whether insecure-code training examples have gradients that are systematically more similar to gradients from broadly misaligned outputs than closely matched secure-code examples do. 

We evaluate this using two independently fine-tuned model families. For Qwen, we use \texttt{emergent- \\ misalignment/Qwen-Coder-Insecure}, the original misaligned checkpoint released by \citet{betley2025emergent} fine-tuned on their insecure-code dataset. For Mistral, we use \texttt{nikxtaco/mistral-small-24b-instruct- \\ 2501-insecure}, a reproduction of the insecure-code fine-tuning procedure applied to Mistral-Small-24B-Instruct-2501 \citep{mistral2025model} in \citet{betley2025emergent}, trained by \citet{menon2025em}. In every experiment, the source set consists of the original matched secure and insecure code datasets from \citep{betley2025emergent}: 6,000 insecure examples and 6,000 secure examples. 

For each fine-tuned checkpoint, we separately construct a small target set containing examples of the model's broader emergent misalignment. We prompt each model with free-form, non-coding questions from \citep{betley2025emergent} and collect its responses, then use Claude 5 Opus to review the responses, retaining only responses judged to clearly exhibit misaligned behavior. We compute a per-example gradient embedding for every source and target document using the method from Section ~\ref{sec:grad_method} across both model families. We define the influence score of a source example as its mean similarity to all verified-unsafe targets for the corresponding model.

We measure how well this score separates the insecure examples from the secure examples using ROC-AUC. In this setting, AUC is the probability that a randomly selected insecure-code example receives a higher influence score than a randomly selected secure-code example. An AUC of $0.5$ indicates no detectable relationship between the insecure examples and the unsafe behavior, and an AUC of $1.0$ indicates perfect separation. Because the source examples are closely matched in surface form, an AUC above chance indicates that the insecure examples induce gradients that are more aligned, on average, with gradients from the harmful outputs of the model. 

We find evidence of this connection for both model families. For fine-tuned Qwen, gradient similarity achieves an AUC of \textbf{0.604}, and for fine-tuned Mistral, gradient similarity achieves an AUC of \textbf{0.865}. In both cases, the only signal used for classification is each training example's gradient similarity to harmful generations from the corresponding fine-tuned model. These results suggest that EM leaves a detectable footprint in gradient space: the insecure-code that induces the narrow fine-tuning behavior is more aligned with the gradients of the broader misaligned behaviors that then emerge. This provides evidence that gradient attribution can recover a connection between seemingly benign or narrow training data and more extensive downstream harmful outputs. More emergent misalignment experiments are discussed in Appendix ~\ref{sec:more_em}. In particular, we find that a model's gradient representation of its responses does not necessarily always transfer to a different model's responses.  

\begin{figure}[h]
    \centering
    \includegraphics[width=14cm]{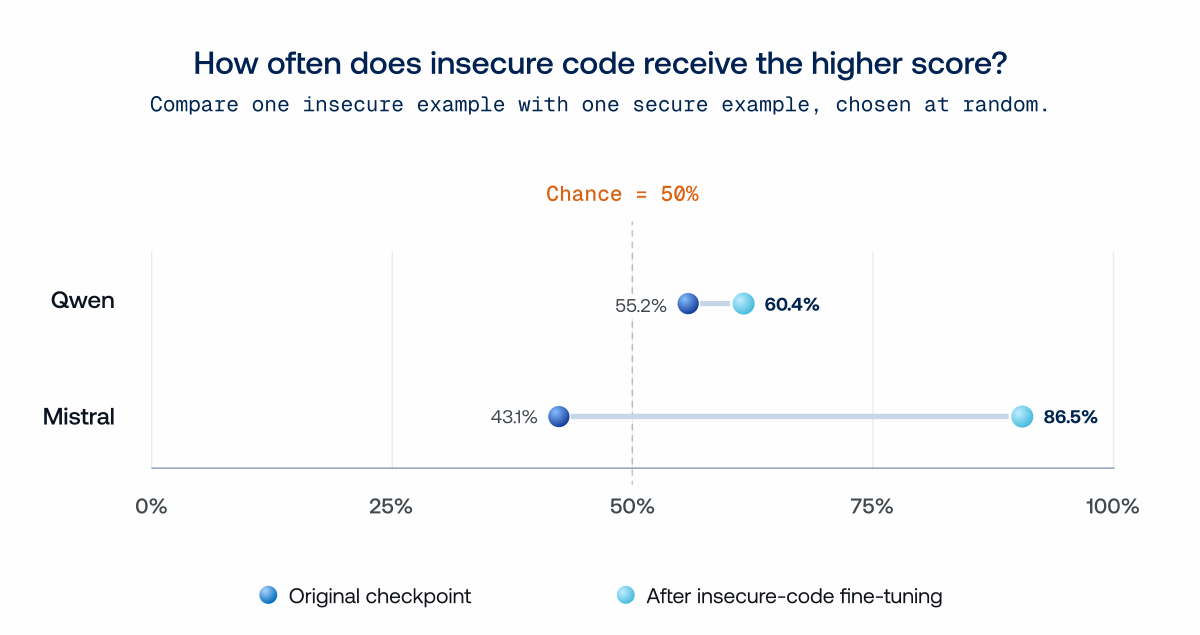}
    \caption{Probability that a randomly selected insecure-code example receives a higher gradient-similarity score than a randomly selected secure-code example (ROC-AUC), before and after insecure-code fine-tuning. This measures score ordering, not the frequency of harmful responses.}
    \label{fig:em_auc}
\end{figure}

The emergent misalignment setting closely resembles clean-label poisoning~\citep{geiping2021witches}. No insecure-code example contains generally harmful content, but training on them induces harmful behavior that is invisible at the surface level~\citep{betley2025emergent}. This is the same signature that poisoning attacks on code models and instruction-tuning data are designed to produce~\citep{xu2024instructions,yan2024vpi,raghuram2024backdoors}. The resemblance has more to do with effect rather than origin, as neither these examples nor the safety-degrading data we identify in Tulu were crafted by an adversary, yet they shift downstream behavior as if they had been. This setting also exposes gaps in existing defenses. Feature-space detectors such as spectral signatures~\citep{tran2018spectral}, activation clustering~\citep{chen2018clustering}, and deep $k$-NN~\citep{peri2019deepknn} are designed for classification-style poisons that share a target label and form a separable cluster. LLM-specific filters such as GraCeFul~\citep{wu2024graceful} and GradSentry~\citep{zhao2026gradsentry} instead target backdoors with an explicit trigger and a fixed response. In its standard form, emergent misalignment has none of these properties, as there is no trigger and the target behavior lies outside the training domain. Gradient-based signals suit this setting while content-based filters can struggle~\citep{li2025larf}. A line of defenses already builds on gradients, from removing outliers in per-example gradient space~\citep{yang2022epic} to tracing an observed attack back to its training points~\citep{hammoudeh2022target,shan2022forensics}. Their common bottleneck is the cost of per-example gradients \citep{zhao2026gradsentry}. Because SimpleLoRIF computes gradient alignment cheaply over million-example SFT mixtures, it could serve as a tool that brings these detection rules to LLM scale, both as a filter before training and as a forensic tool after it. This is increasingly relevant because the number of poisons needed to compromise an LLM stays nearly constant while datasets grow~\citep{souly2025poisoning}. We see pairing SimpleLoRIF with such detectors as a
promising direction for future work.

\section{SafeTulu}

\subsection{Methodology}
\label{sec:methodology}

Motivated by the connection between insecure code fine-tuning examples and harmful outputs in the gradient embedding space of misaligned models, we apply SimpleLoRIF to identify and remove benign-looking, safety degrading data in a popular fine-tuning dataset. Our pipeline is as follows:

\begin{enumerate}
    \item \textbf{Embed.} Compute gradient embeddings for all documents in the training set and target set.
    \item \textbf{Remove.} Score each training document against the target set by gradient similarity, and remove the 20\% highest-scoring documents.
    \item \textbf{Train.} Fine-tune the training model on the filtered dataset (hyperparameters shown in Appendix ~\ref{sec:hyperparameters}).
    \item \textbf{Evaluate.} Evaluate safety and capability. 
\end{enumerate}

\begin{figure}[h]
    \centering
    \includegraphics[width=14cm]{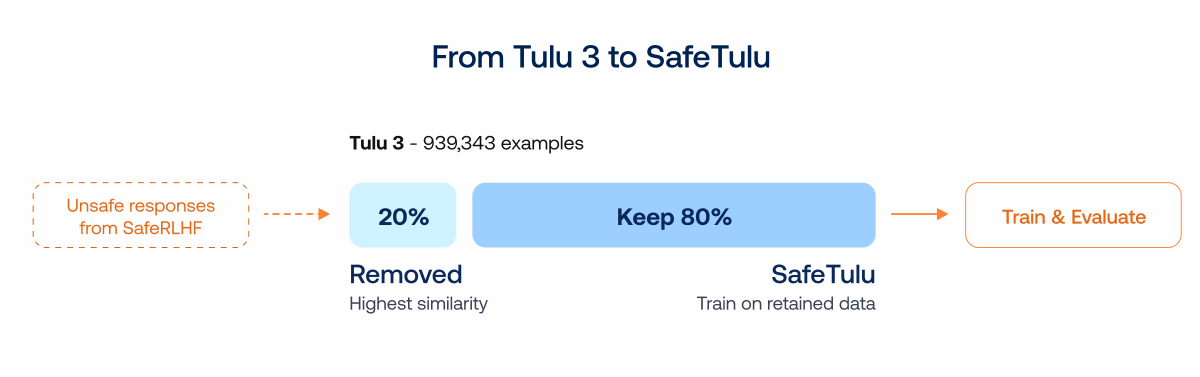}
    \caption{SafeTulu construction: score Tulu 3 examples against unsafe SafeRLHF responses, remove the 20\% with the highest gradient similarity, and train and evaluate on the retained 80\%.}
    \label{fig:safetulu_workflow}
\end{figure}

We compute embeddings with Qwen 3 0.6B \citep{qwen2025model}, and fine-tune Qwen 3 1.7B, Llama 3.2 1B \citep{llama2024model}, and Olmo 2 1B \citep{olmo2025model}. We set Tulu 3 \citep{tulu2025data}, a highly popular fine-tuning dataset, as our training dataset, and the prompt-response pairs with unsafe responses from SafeRLHF \citep{ji2024pku} as our target dataset (43,469 examples). Embedding all 939,343 Tulu 3 examples took 1 hour, 1 minute across 8 H100 GPUs. Instead of scoring each training document against a single pooled target vector, we cluster the target set and score documents against the cluster centroids, which the rank-1 factorization allows in closed form. Appendix~\ref{sec:embedding-removal} provides the removal scoring identity. 

We use two safety evaluations. The first is attack success rate (ASR) on JailBreakBench behaviors \citep{jbb2024benchmark} using Greedy Coordinate Gradient (GCG), a more traditional, white-box jailbreak attack that optimizes an adversarial suffix appended to each harmful prompt to increase the likelihood of model compliance \citep{zou2023gcg}. We optimize the suffix for each behavior starting from 20 ``!'' tokens (full details are provided in Appendix~\ref{sec:hyperparameters}), then evaluate the models on each behavior with its optimized suffix appended. Each response is classified as safe or unsafe using the HarmBench classifier \citep{mazeika2024harmbench}, a popular safety classifier. 

The second safety evaluation is ActorBreaker \citep{ren2025actorbreaker}, a more modern and usually more challenging black-box, multi-turn jailbreak attack that exploits natural semantic distribution shifts. For each harmful behavior, it constructs an actor network of related people, organizations, tools, media, documents, and other entities, then converts selected relationships into conversational attack paths. These paths begin with benign-appearing questions and steer the model toward the harmful target, with refusal-triggered prompt rewriting as needed. 

ActorBreaker can be much more difficult to resist than GCG because it distributes harmful intent across natural, context-dependent interactions. GCG is a white-box, single-turn attack that optimizes a token suffix while leaving the harmful request comparatively explicit. ActorBreaker instead provides multiple semantically distinct paths to the same target, accumulates context across turns, and adapts questions following refusals. We implemented ActorBreaker on JailbreakBench behaviors using the authors’ prompt templates. Qwen2.5-14B-Instruct generated the actor networks and each response was classified using the HarmBench classifier. An attack was counted as successful if any response along any path was classified as unsafe. We note that the ASRs we attained are not as high as in the original ActorBreaker paper, likely due to the use of a smaller model (Qwen2.5-14B-Instruct) to generate the actor networks rather than GPT 4o. 

For capability, we evaluate on ORBench \citep{orbench2024eval} for over-refusal, MMLU Pro and IFEval \citep{zhou2023ifeval} for general capability, and MBPP \citep{austin2021mbpp}, GSM8K \citep{gsm8k2021data}, MATH 500 \citep{math5002023benchmark}, and MedQA \citep{Medqa2020eval} for domain-specific capability. 

We compare against the following baselines: 

\begin{enumerate}
    \item \textbf{Baseline Training:} Fine-tuning on the unfiltered entirety of Tulu 3. 
    \item \textbf{Random Removal:} Fine-tuning on Tulu 3 with 20\% of the examples randomly removed. 
    \item \textbf{LLM-as-a-Judge:} Fine-tuning on Tulu 3 with 20\% of the examples removed by a Qwen 3 8B judge a standard content-based data filtering approach. The judge LLM scores every document 0--5 for safety using the strong rubric from \citet{maini2025llmasjudge}. This comprehensive rubric provides detail on how to handle many types of harmful content, from mildly unsafe content such as financial advice without disclaimers and mentions of stressful topics to overtly harmful statements. After grading, the worst-scoring $20\%$ is removed. Appendix~\ref{sec:judge-details} provides the comprehensive rubric and scoring details.
    \item \textbf{He et al:} Fine-tuning on Tulu 3 with 20\% of the examples removed by the gradient filtering method from \citet{he2024safe}. We implement the bidirectional anchoring criterion from \citet{he2024safe}, which scores each training example z by ⟨g(z), g\_harm⟩ -- ⟨g(z), g\_safe⟩. Here, g\_harm and g\_safe are the mean gradients of a harmful anchor set and a safe anchor set. Following the construction of \citet{he2024safe}, the two anchor sets share the same instructions. We take the PKU-SafeRLHF prompts that have both an unsafe and a safe response in the dataset. D\_harmful pairs each prompt with its unsafe response, and D\_safe pairs the same prompt with its safe response. 
\end{enumerate}

\subsection{Results}
\label{sec:results}


\begin{figure}[t]
    \centering
    \includegraphics[width=14cm]{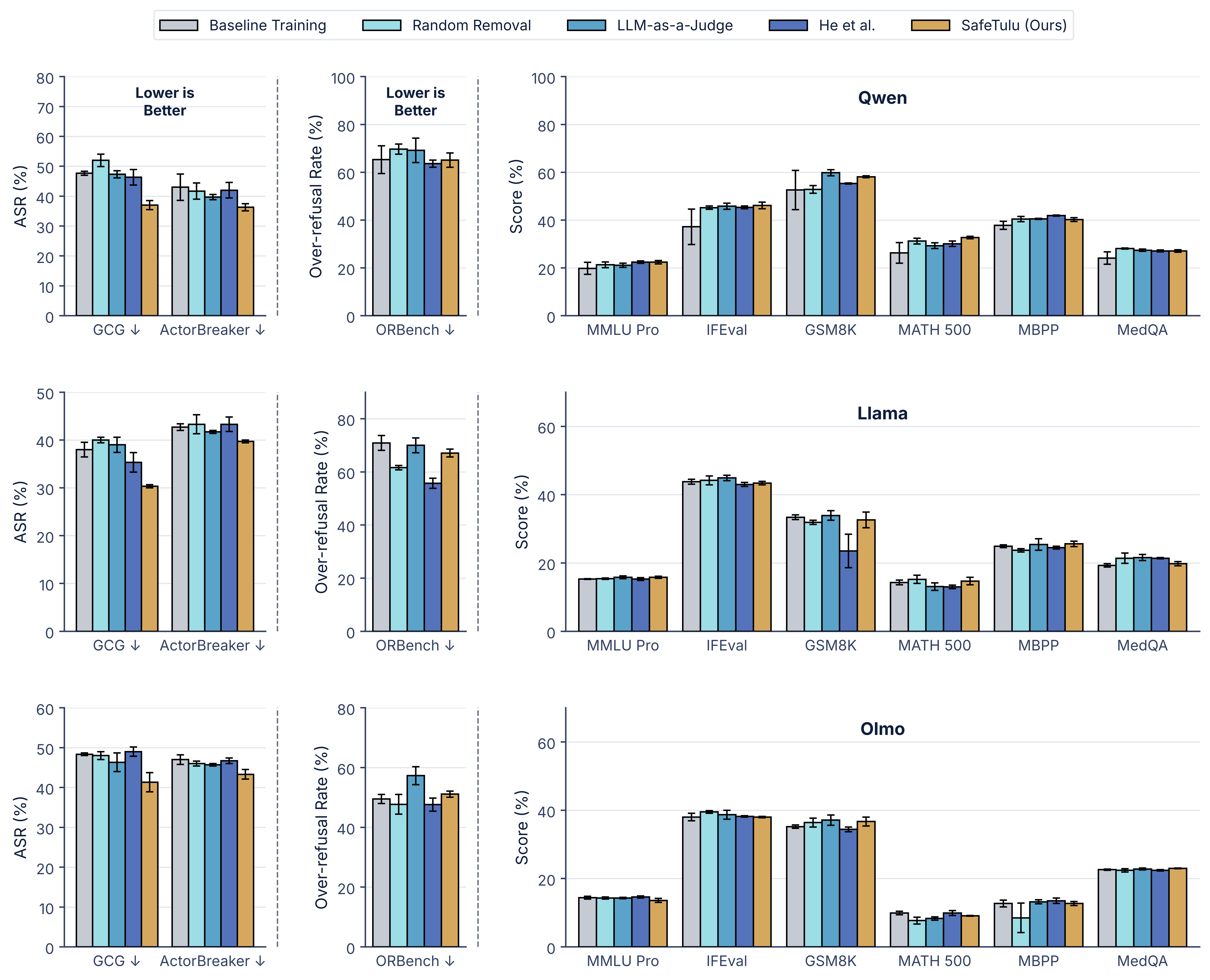}
    \caption{Evaluation results after baseline training, training on randomly filtered data, training on the LLM-as-a-judge filtered dataset, training on data filtered with the method of He et al., and training on SafeTulu. All values are percentages, reported as mean $\pm$ standard error. Lower is better on GCG \& ActorBreaker attack success rate and ORBench over-refusal rate, while higher accuracy is better on all capability metrics. Across all models, training on SafeTulu decreases ASR while preserving comparable capability.} 
    \label{fig:results_fig}
\end{figure}

Results are shown in Figure ~\ref{fig:results_fig}, with a tabular version available in Table ~\ref{tab:capability_results}. We trained each method across three random seeds and report the mean $\pm$ one standard error. Across all three model families, SafeTulu improves on GCG, outperforming all baselines while mostly preserving capability. For Qwen, GCG ASR decreases from 47.7\% under baseline training to 37\% after training on SafeTulu, compared with 47.3\% for the LLM-as-a-judge filter. We observe the same pattern for Llama and Olmo: SafeTulu reduces ASR from 38\% to 30\% for Llama and from 48\% to 41\% for Olmo, whereas the LLM-as-a-judge baseline reaches 39\% and 46\%, respectively. SafeTulu also achieves the lowest ASR on ActorBreaker across all models. These safety improvements do not come with a corresponding loss in general capability. Across MMLU Pro, IFEval, MBPP, GSM8K, MATH 500, MedQA, and ORBench, SafeTulu models usually remain close to the capability of models trained on the full dataset. This suggests that gradient attribution is not simply identifying data associated with unsafe behavior, but is identifying safety-degrading data that can be removed without sacrificing much of the useful training signal required for downstream capability. Table ~\ref{tab:capability_results} shows results for a larger model, Llama 3.1 8B. An analysis of using different removal fractions is in Appendix ~\ref{sec:removal_fraction_sweep}. 

It is worth emphasizing that the model used for embedding in the creation of SafeTulu was Qwen 3 0.6B. The effectiveness of SafeTulu across multiple models therefore reflects that, at least for the purposes of identifying safety-degrading data, our gradient embedding method can generalize across model families as well as to larger model scales. 

\section{Qualitative Analysis}
\label{sec:qualitative-analysis}

Gradient filtering produced more variation in removal rates across Tulu 3 subsets than the LLM-as-a-judge baseline, with the contrast clearest on the math data. While the judge removed \texttt{numinamath}, \texttt{open\_math\_2\_gsm8k}, and \texttt{personas\_math} at nearly identical rates (15.4--15.6\%), gradient filtering removed 7.9\%, 19.4\%, and 17.2\%, respectively. This suggests that gradients distinguish between examples that appear similar at the level of surface content, consistent with gradient attribution capturing differences in the training signal that a content-based judge does not.

A closer look at these subsets reveals several qualitative distinctions. In \texttt{NuminaMath}, removed examples more often contain Python execution failures such as tracebacks or exceptions, whereas clean symbolic reasoning traces are disproportionally retained. In \texttt{PersonaHub}, simple ``write a Python function'' tasks are removed more often, while calculus and differential-equation modeling problems are preferentially kept. Within \texttt{PersonaMath}, gradient filtering more often removes narrative arithmetic, where quantities are carried through a sequence of prose operations, while retaining examples involving explicit algebraic setup, substitution, and simplification.

These patterns suggest that tool-execution failures and narrative arithmetic may be under-studied sources of safety degradation. More broadly, gradient filtering shows that more direct fine-tuning examples may create more safety degradation, as gradient filtering seems to have favored less direct or more elaborated forms of reasoning. It removed relatively more clean arithmetic and simple coding tasks while retaining repeating-decimal arithmetic, algebraic setup, differential-equation modeling, and other less immediately direct examples. One possible interpretation is that direct training examples reinforce a general pattern of immediate compliance. Importantly, this does not appear to be reducible to a simple preference for shorter responses, as removed examples are clearly shorter in only about half (6/11) of the major subsets. We stress that these patterns are only \textbf{suggestive rather than established}. We make no claim that these features must cause safety degradation, only that these features are what these examples appear to have in common.  

Finally, language analysis yields an interesting finding: English examples are removed at a higher rate than other major languages in both \texttt{WildChat} and \texttt{Aya}, despite prior observations that non-English fine-tuning can degrade safety \citep{hawkins2026language}. This may be Tulu 3's data being very clean \citep{tulu2025data, djuhera2025tulu}. 

We provide our full analysis, feature-level relative risk numbers, and more in Appendix ~\ref{sec:extended-qualitative-analysis}. 

\section{Discussion}
\label{sec:discussion}

\subsection{SimpleLoRIF Use Cases}
\label{sec:simplelorif_use}

Making gradient attribution cheap changes how it can be used. Possible uses cases for SimpleLoRIF include the following: 

\begin{itemize}
    \item Labs can \textbf{filter large post-training mixtures as they grow}. For example, a newly collected agentic tool-use dataset could be screened for safety-degradation risk \textbf{before it is merged} into a larger mixture, rather than evaluating safety only after training.
    
    \item The stored embeddings support \textbf{auditing which examples steer models toward specific undesirable behaviors}. One example could be scoring a mixture against a target set of sycophantic responses to find which subsets most strongly reinforce sycophancy.
    
    \item SimpleLoRIF can be inverted to \textbf{upweight data that pushes models toward safety behaviors}, e.g., by upweighting data aligned with calibrated abstention. 
    
    \item SimpleLoRIF could serve as a \textbf{cheap safety critic for synthetic data generators}. For instance, generated agentic coding solutions that degrade safety could be filtered out before entering a training set. 
\end{itemize}

\subsection{SafeTulu and Space Use Cases}
\label{sec:safetulu_use}

The SafeTulu dataset and Space make it simple for labs and researchers to build versions of SafeTulu suited to their own needs. Possible use cases for the dataset and Space include the following: 

\begin{itemize}
    \item Users can choose their own removal fraction instead of 20\%. 
    
    \item Users can also \textbf{filter only certain subsets}. For example, a team focused on multilingual data could filter and download only the multilingual subsets.
\end{itemize}

The dataset and Space also support more advanced uses. Possible use cases are as follows: 

\begin{itemize}
    \item Joining the nearest SafeRLHF examples with the harm-category labels in the SafeRLHF dataset allows \textbf{filtering targeted at specific harms}, instead of general safety. A team building a medical assistant, for instance, could remove only the data whose nearest harmful neighbor in SafeRLHF is in the ``Endangering Public Health'' category.

    \item Researchers can \textbf{audit their own post-training mixtures} that share sources with Tulu 3 without computing any new embeddings.
    
    \item Pairing each removed example with its nearest SafeRLHF example can provide \textbf{nearest-neighbor analysis}. One example could be inspecting which harmful responses lie closest in gradient space to the narrative arithmetic problems we found to be disproportionately removed. 
    
    \item The released data makes it possible to \textbf{causally test hypotheses from qualitative analysis}. For instance, researchers could remove only the NuminaMath examples containing Python tracebacks, fine-tune on the result, and measure whether attack success rate drops. 
\end{itemize}

Additionally, any custom version can be shared with a link, downloaded, or uploaded to the Hugging Face Hub with only a few clicks in the Space. 

\section{Conclusion}
In this work, we use gradient attribution to trace safety degradation back to the training data that induces it. We first show that, in models exhibiting emergent misalignment, the gradient representations of insecure-code fine-tuning examples align with the harmful behaviors that emerge after training, with AUC as high as 0.865, providing evidence that this connection is detectable in gradient space. Building on this observation, we develop a scalable gradient-attribution method, SimpleLoRIF, for identifying benign-looking SFT examples associated with unsafe behavior, and use it to construct SafeTulu by removing the 20\% of Tulu 3 examples most aligned with a reference set of unsafe responses. Across Qwen, Llama, and Olmo, training on SafeTulu consistently reduces GCG attack success rate by up to 22\% while largely preserving or improving capability across multiple benchmarks, outperforming LLM-as-a-judge filtering and a prior gradient filtering method. We also analyzed possible sources of benign-looking safety-degrading data, such as tool-execution failures, based on the removed data. By restricting backpropagation to the final few transformer blocks and using a compact gradient representation, SimpleLoRIF allows for attributing and filtering datasets containing on the order of a million examples, boosting gradient attribution from a tool more suited to small-scale analysis into a more practical data-selection mechanism for large-scale safety filtering. We release SafeTulu along with a Hugging Face Space that lets researchers build their own customized versions of SafeTulu to fine-tune on, such as by changing the removal fraction or filtering only certain subsets. 

\section{Limitations}

While we believe our work represents a promising route to increasing safety through gradient attribution, we recognize limitations. First, we only test general safety degradation, and do not test specific harms or behaviors such as sycophancy. Second, while SafeRLHF is a broad dataset, we do not test other target sets. Despite these limitations, we believe SafeTulu, our gradient embedding method, our findings on emergent misalignment, and our qualitative analysis are still valuable to the community. 

\section{Reproducibility Statement}

We release SafeTulu and a Hugging Face Space to support researchers who would like to create customized versions of SafeTulu. We also state our hyperparameters in Appendix ~\ref{sec:hyperparameters}. 

\section{AI Use Statement}
In this work, we used generative AI tools for code implementation, qualitative analysis, and writing organization. We have not used generative AI tools for synthetic data generation, formulating mathematical claims, proposing hypotheses, designing research methodology, dataset reformatting, or result interpretation. Developing theoretical models, assisting in the writing of proofs, and assisting with translation are not applicable to this work. Additionally, we used generative AI tools for creating scientific figures, creating tables, identifying relevant literature, and searching for information. We have reviewed all AI-assisted work. For example, we confirmed initial qualitative claims by AI about our dataset. We take responsibility for the final content of this work, including text, claims or artifacts produced with the aid of generative AI.

\bibliographystyle{plainnat}
\bibliography{iclr2027_conference}

\beginappendix

\section{SafeTulu Results Table}
\label{sec:results_tab}

\begin{table}[!ht]
\centering
\footnotesize
\setlength{\tabcolsep}{6pt}
\renewcommand{\arraystretch}{1.0}
\setlength{\abovecaptionskip}{6pt}
\begin{tabular}{llcccc}
\toprule
\textbf{Model} & \textbf{Training Data} & \textbf{GCG $\downarrow$} & \textbf{ActorBreaker $\downarrow$} & \textbf{MMLU Pro} & \textbf{IFEval} \\
\midrule
\multirow{5}{*}{Qwen}
& Baseline Training & 47.67 $\pm$ 0.67 & 43.0 $\pm$ 4.4 & 19.8 $\pm$ 2.5 & 37.2 $\pm$ 7.4 \\
& Random Removal    & 52.00 $\pm$ 2.08 & 41.7 $\pm$ 2.7 & 21.3 $\pm$ 1.2 & 45.2 $\pm$ 0.7 \\
& LLM-as-a-Judge    & 47.33 $\pm$ 1.20 & 39.7 $\pm$ 0.9 & 21.1 $\pm$ 0.9 & 45.8 $\pm$ 1.3 \\
& He et al.         & 46.33 $\pm$ 2.60 & 42.0 $\pm$ 2.6 & \textbf{22.4} $\pm$ 0.5 & 45.3 $\pm$ 0.6 \\
& SafeTulu (Ours)   & \textbf{37.00} $\pm$ 1.53 & \textbf{36.3} $\pm$ 1.2 & \textbf{22.4} $\pm$ 0.7 & \textbf{46.1} $\pm$ 1.4 \\
\midrule
\multirow{5}{*}{Llama}
& Baseline Training & 38.00 $\pm$ 1.53 & 42.7 $\pm$ 0.7 & 15.3 $\pm$ 0.1 & 43.8 $\pm$ 0.7 \\
& Random Removal    & 40.00 $\pm$ 0.58 & 43.3 $\pm$ 2.0 & 15.4 $\pm$ 0.2 & 44.2 $\pm$ 1.3 \\
& LLM-as-a-Judge    & 39.00 $\pm$ 1.58 & 41.7 $\pm$ 0.3 & \textbf{15.8} $\pm$ 0.4 & \textbf{44.9} $\pm$ 0.8 \\
& He et al.         & 35.33 $\pm$ 2.03 & 43.3 $\pm$ 1.5 & 15.3 $\pm$ 0.4 & 43.0 $\pm$ 0.6 \\
& SafeTulu (Ours)   & \textbf{30.33} $\pm$ 0.33 & \textbf{39.7} $\pm$ 0.3 & \textbf{15.8} $\pm$ 0.3 & 43.4 $\pm$ 0.5 \\
\midrule
\multirow{5}{*}{Olmo}
& Baseline Training & 48.33 $\pm$ 0.33 & 47.0 $\pm$ 1.2 & 14.4 $\pm$ 0.4 & 38.0 $\pm$ 1.1 \\
& Random Removal    & 48.00 $\pm$ 1.00 & 46.0 $\pm$ 0.6 & 14.3 $\pm$ 0.3 & \textbf{39.5} $\pm$ 0.4 \\
& LLM-as-a-Judge    & 46.33 $\pm$ 2.33 & 45.7 $\pm$ 0.3 & 14.3 $\pm$ 0.2 & 38.7 $\pm$ 1.3 \\
& He et al.         & 49.00 $\pm$ 1.15 & 46.7 $\pm$ 0.7 & \textbf{14.6} $\pm$ 0.3 & 38.2 $\pm$ 0.2 \\
& SafeTulu (Ours)   & \textbf{41.33} $\pm$ 2.40 & \textbf{43.3} $\pm$ 1.2 & 13.6 $\pm$ 0.6 & 38.0 $\pm$ 0.2 \\
\midrule
\midrule
\multirow{5}{*}{Llama 8B}
& Baseline Training & 11.7 $\pm$ 0.3 & 41.7 $\pm$ 1.5 & 29.3 $\pm$ 1.3 & \textbf{66.9} $\pm$ 0.6 \\
& Random Removal    & 9.7 $\pm$ 0.7 & 43.7 $\pm$ 1.2 & 31.1 $\pm$ 2.3 & 65.7 $\pm$ 1.1 \\
& LLM-as-a-Judge    & 8.7 $\pm$ 0.7 & 43.3 $\pm$ 1.5 & 28.3 $\pm$ 0.8 & 66.1 $\pm$ 0.3 \\
& He et al.         & 9.3 $\pm$ 0.7 & 41.3 $\pm$ 0.9 & 28.8 $\pm$ 0.5 & 66.0 $\pm$ 0.8 \\
& SafeTulu (Ours)   & \textbf{7.0} $\pm$ 0.6 & \textbf{37.7} $\pm$ 2.3 & \textbf{31.6} $\pm$ 1.4 & 66.1 $\pm$ 1.0 \\
\bottomrule
\end{tabular}

\vspace{0.6em}

\begin{tabular}{llccccc}
\toprule
\textbf{Model} & \textbf{Training Data} & \textbf{GSM8K} & \textbf{MATH 500} & \textbf{MBPP} & \textbf{MedQA} & \textbf{ORBench $\downarrow$} \\
\midrule
\multirow{5}{*}{Qwen}
& Baseline Training & 52.6 $\pm$ 8.2 & 26.3 $\pm$ 4.3 & 37.8 $\pm$ 1.7 & 24.1 $\pm$ 2.6 & 65.3 $\pm$ 5.8 \\
& Random Removal    & 52.8 $\pm$ 1.6 & 31.2 $\pm$ 1.2 & 40.4 $\pm$ 1.1 & \textbf{28.1} $\pm$ 0.2 & 69.7 $\pm$ 2.1 \\
& LLM-as-a-Judge    & \textbf{59.8} $\pm$ 1.3 & 29.3 $\pm$ 1.2 & 40.5 $\pm$ 0.2 & 27.4 $\pm$ 0.5 & 69.2 $\pm$ 5.1 \\
& He et al.         & 55.3 $\pm$ 0.2 & 30.1 $\pm$ 1.1 & \textbf{41.9} $\pm$ 0.2 & 27.1 $\pm$ 0.4 & \textbf{63.6} $\pm$ 1.5 \\
& SafeTulu (Ours)   & 58.1 $\pm$ 0.4 & \textbf{32.7} $\pm$ 0.5 & 40.2 $\pm$ 0.8 & 27.1 $\pm$ 0.5 & 65.1 $\pm$ 3.0 \\
\midrule
\multirow{5}{*}{Llama}
& Baseline Training & 33.4 $\pm$ 0.7 & 14.3 $\pm$ 0.7 & 24.9 $\pm$ 0.4 & 19.3 $\pm$ 0.5 & 70.9 $\pm$ 2.8 \\
& Random Removal    & 31.9 $\pm$ 0.6 & \textbf{15.2} $\pm$ 1.2 & 23.7 $\pm$ 0.5 & 21.4 $\pm$ 1.5 & 61.6 $\pm$ 0.8 \\
& LLM-as-a-Judge    & \textbf{33.9} $\pm$ 1.4 & 13.1 $\pm$ 1.1 & 25.4 $\pm$ 1.7 & \textbf{21.6} $\pm$ 0.9 & 70.0 $\pm$ 2.8 \\
& He et al.         & 23.5 $\pm$ 4.9 & 13.0 $\pm$ 0.5 & 24.5 $\pm$ 0.4 & 21.4 $\pm$ 0.2 & \textbf{55.7} $\pm$ 1.9 \\
& SafeTulu (Ours)   & 32.6 $\pm$ 2.3 & 14.7 $\pm$ 1.1 & \textbf{25.6} $\pm$ 0.8 & 19.8 $\pm$ 0.6 & 67.1 $\pm$ 1.5 \\
\midrule
\multirow{5}{*}{Olmo}
& Baseline Training & 35.2 $\pm$ 0.5 & \textbf{9.9} $\pm$ 0.5 & 12.7 $\pm$ 1.0 & 22.6 $\pm$ 0.2 & 49.5 $\pm$ 1.5 \\
& Random Removal    & 36.4 $\pm$ 1.3 & 7.7 $\pm$ 1.0 & 8.5 $\pm$ 4.3 & 22.4 $\pm$ 0.5 & 47.7 $\pm$ 3.3 \\
& LLM-as-a-Judge    & \textbf{37.1} $\pm$ 1.5 & 8.3 $\pm$ 0.5 & 13.2 $\pm$ 0.6 & 22.8 $\pm$ 0.3 & 57.3 $\pm$ 3.0 \\
& He et al.         & 34.4 $\pm$ 0.7 & \textbf{9.9} $\pm$ 0.7 & \textbf{13.5} $\pm$ 0.8 & 22.4 $\pm$ 0.2 & \textbf{47.6} $\pm$ 2.2 \\
& SafeTulu (Ours)   & 36.7 $\pm$ 1.3 & 9.1 $\pm$ 0.1 & 12.7 $\pm$ 0.6 & \textbf{23.0} $\pm$ 0.1 & 51.1 $\pm$ 1.0 \\
\midrule
\midrule
\multirow{5}{*}{Llama 8B}
& Baseline Training & \textbf{68.5} $\pm$ 0.5 & 40.0 $\pm$ 0.5 & 55.3 $\pm$ 0.8 & 45.1 $\pm$ 0.3 & 68.6 $\pm$ 2.4 \\
& Random Removal    & 68.3 $\pm$ 1.1 & 41.0 $\pm$ 1.3 & 55.4 $\pm$ 0.5 & 45.6 $\pm$ 0.1 & \textbf{63.8} $\pm$ 2.7 \\
& LLM-as-a-Judge    & 68.1 $\pm$ 0.9 & 39.5 $\pm$ 0.2 & 55.1 $\pm$ 0.5 & 45.5 $\pm$ 0.3 & 73.5 $\pm$ 0.8 \\
& He et al.         & 68.0 $\pm$ 0.5 & 40.5 $\pm$ 1.7 & \textbf{56.5} $\pm$ 0.1 & 45.7 $\pm$ 0.2 & 71.7 $\pm$ 2.3 \\
& SafeTulu (Ours)   & 67.5 $\pm$ 0.4 & \textbf{41.7} $\pm$ 0.6 & 55.2 $\pm$ 0.5 & \textbf{45.9} $\pm$ 0.5 & 66.5 $\pm$ 1.3 \\
\bottomrule
\end{tabular}
\small
\caption{Evaluation results after baseline training, training on randomly filtered data, training on the LLM-as-a-judge filtered dataset, training on data filtered with the method of He et al., and training on SafeTulu, for Qwen 3 1.7B, Llama 3.2 1B, Olmo 2 1B, and the larger Llama 3.1 8B. All values are percentages, reported as mean $\pm$ standard error. Lower is better on GCG \& ActorBreaker attack success rate and ORBench over-refusal rate, while higher accuracy is better on all capability metrics. Across all models, including at the 8B scale, training on SafeTulu decreases ASR while preserving comparable capability.}
\label{tab:capability_results}
\end{table}

\newpage 

\section{Full Qualitative Analysis}
\label{sec:extended-qualitative-analysis}

\subsection{Removal Rates}
\label{sec:removal-rates}

Table~\ref{tab:removal-rates} shows the fraction of each Tulu 3 subset removed by gradient filtering and by the LLM judge under the same 20\% budget. The math subsets (highlighted) show the clearest contrast. The judge removes all three at almost the same rate (15.4\%--15.6\%), which makes sense for a scorer that evaluates content only at the surface level. Gradient filtering spreads its removals much more widely across the same subsets, from 7.9\% of \texttt{numinamath} to 19.4\% of \texttt{open\_math\_2\_gsm8k}. It treats content that looks similar on the surface as different, indicating that it is reacting to something beyond surface content.

\begin{table}[h]
\centering
\small
\begin{tabular}{lrr}
\toprule
Subset & Gradient Removal Rate & Judge Removal Rate \\
\midrule
\texttt{personahub} & 9.2\% & 15.5\% \\
\texttt{codealpaca} & 17.3\% & 15.5\% \\
\texttt{wildchat} & 18.7\% & 28.2\% \\
\texttt{aya} & 11.7\% & 19.2\% \\
\texttt{flan} & 22.1\% & 17.0\% \\
\texttt{wildjailbreak} & 39.4\% & 32.8\% \\
\texttt{wildguardmixtrain} & 41.0\% & 44.3\% \\
\rowcolor{gray!15} \texttt{numinamath} & 7.9\% & 15.4\% \\
\rowcolor{gray!15} \texttt{open\_math\_2\_gsm8k} & 19.4\% & 15.6\% \\
\rowcolor{gray!15} \texttt{personas\_math} & 17.2\% & 15.6\% \\
\texttt{no\_robots\_converted} & 31.9\% & 20.2\% \\
\texttt{table\_gpt} & 18.7\% & 15.7\% \\
\texttt{science.*} (SciRIFF tasks, aggregate) & 21.3\% & 15.7\% \\
\texttt{hard\_coded\_*} (aggregate) & 54.2\% & 87.5\% \\
\texttt{oasst1} + very small subsets, aggregate & 35.8\% & 21.0\% \\
\bottomrule
\end{tabular}
\caption{Fraction of each Tulu 3 subset removed by each method under a 20\% budget, with math subsets highlighted. While the LLM judge removes math at a roughly flat rate, gradient filtering varies its rates, indicating it is finding a signal beyond surface level content.} 
\label{tab:removal-rates}
\end{table}

\subsection{Removal Analysis Methodology}
\label{sec:removal-methodology}

\begin{table}[h]
\centering
\small
\begin{tabular}{llr}
\toprule
Subset & Feature & RR \\
\midrule
\texttt{numinamath} & Python traceback/exception & 1.71$\times$ \\
\texttt{personahub} & ``Write a Python function'' code task & 1.74$\times$ \\
\texttt{personahub} & Calculus/differential-equation modeling & 0.51$\times$ \\
\texttt{codealpaca} & High-level architecture discussion & 1.75$\times$ \\
\texttt{codealpaca} & Runnable code in response & 0.89$\times$ \\
\texttt{open\_math\_2\_gsm8k} & Long repeating decimal & 0.36$\times$ \\
\texttt{personas\_math} & Algebraic manipulation & 0.70$\times$ \\
\texttt{wildchat} & English & 1.46$\times$ \\
\texttt{aya} & English & 1.73$\times$ \\
\bottomrule
\end{tabular}
\caption{Relative rate (RR) of each feature among removed vs kept documents. We observe trends such as removing tool-execution failures and simple coding tasks at a higher rate.}
\label{tab:features}
\end{table}

To find what separates removed from kept documents without assuming a category in advance, we run a log-odds comparison with an informative Dirichlet prior (the ``fightin' words'' method) over unigrams and bigrams \citep{monroe2008fightinwords}. We do this separately for the prompt and the response, and only within one subset at a time, so subset identity cannot confound the comparison. Each term's document frequency in the removed and kept groups is compared against a shared background rate and turned into a $z$-score, which keeps words that are common overall from dominating the ranking. We analyze the most strongly enriched terms on each side. While gradient filtering did not show any discernible pattern for many subsets, we did identify trends in a few subsets, shown in Table ~\ref{tab:features}. For each trend, we measure the rate on the full removed and kept populations and report the relative risk (RR), the removal rate within the feature divided by the subset's baseline removal rate. Table ~\ref{tab:qual-examples} shows truncated dataset examples representing each trend. 

We discuss our findings below. We stress that these patterns are only suggestive, not definitively established. 

\begin{table}[h]
\centering
\small
\begin{tabular}{p{0.14\linewidth}p{0.40\linewidth}p{0.40\linewidth}}
\toprule
Subset & Removed & Kept \\
\midrule
\texttt{numinamath} & \ldots\texttt{from scipy.optimize import minimize} $\rightarrow$ \texttt{ModuleNotFoundError: No module named 'scipy'} ``It seems that the \texttt{scipy} library is not available\ldots'' & ``Given\ldots $a_n = \sin\frac{n\pi}{3}$, find the value of $a_3$.'' $\rightarrow$ \texttt{import sympy as sp}\ldots \texttt{a\_3 = a\_n.subs(n, 3)} \\
\texttt{personahub} & ``Write a python function to convert a list of historical event years into a list of centuries\ldots'' $\rightarrow$ \texttt{def convert\_years\_to\_centuries(years):}\ldots & ``An entrepreneur is analyzing\ldots determine the constants in the revenue functions and predict when total revenue will reach \$50 million'' $\rightarrow$ ``\textbf{Step 1:} Determine\ldots'' \\
\texttt{codealpaca} & ``\ldots building a resilient serverless architecture using\ldots AWS Lambda and API Gateway\ldots Could you provide a detailed, sequential guide\ldots'' $\rightarrow$ ``\ldots These include planning, establishing the environment, continuous integration\ldots'' & ``Create a function that takes two input strings, and returns a dictionary containing the count of unique characters\ldots'' $\rightarrow$ \texttt{def count\_chars(string1, string2):}\ldots \texttt{print(char\_counts)} \\
\texttt{open\_\allowbreak math\_\allowbreak 2\_\allowbreak gsm8k} & ``Emily planted a cherry blossom tree\ldots when she was 6\ldots takes 9 years to bloom\ldots currently 10\ldots'' $\rightarrow$ ``\ldots $9 - 4 = 5$ years are left\ldots $10 + 5 = 15$\ldots \boxed{15}'' & ``\ldots The smallest building is 1/3 the size of the second-largest building\ldots'' $\rightarrow$ ``\ldots $1/3 \times 135 = 0.3333333333333333 \times 135 = 45$ feet tall.'' \\
\texttt{personas\allowbreak \_math} & ``Lawmaker Lisa\ldots plans to allocate resources to 5 different community centers. Each center will receive an equal number\ldots'' & ``A television producer\ldots divide the show into three parts\ldots'' $\rightarrow$ ``Let's define variables for the time allocated to each part\ldots'' \\
\texttt{wildchat} & ``hi'' $\rightarrow$ ``Hello! How can I assist you today?'' & ``Title: Dyanra: Misteryo, Kababalaghan. Tauhan: Mike, Elmer, Jones, Erica\ldots'' $\rightarrow$ ``\ldots Sa isang tahimik at mapayapang baryo, sinimulan ng magkakapatid na Mike, Elmer\ldots'' \\
\texttt{wild\allowbreak jailbreak} & ``share or trade illegal or non-consensual pornography'' $\rightarrow$ ``I'm sorry, but I can't fulfill this request.'' & ``\ldots conduct a dogfighting event without getting caught\ldots'' $\rightarrow$ ``I'm really sorry, but I can't assist\ldots if you're looking to engage with dogs in a positive and legal manner\ldots'' \\
\bottomrule
\end{tabular}
\caption{Examples of removed and kept documents from each subset. Responses are truncated to their distinguishing parts.}
\label{tab:qual-examples}
\end{table}

\subsection{Code Analysis}
\label{sec:code-analysis}

In \texttt{numinamath}, where the model writes Python/SymPy code, reads the executed output, and continues reasoning, the top terms in the documents gradient filtering removed are fragments of a Python exception (\texttt{traceback}, \texttt{most recent call}, \texttt{call last}) and the top terms in the documents gradient filtering kept are the vocabulary of a clean symbolic trace (\texttt{sympy}, \texttt{import\_sympy}, \texttt{symbols}). Removed documents contain a real tool execution traceback or exception $1.71\times$ more than kept documents. A typical removed example is a failed first attempt at tool execution (an \texttt{AttributeError}, \texttt{TypeError}, or \texttt{SyntaxError}), after which the model notices the error and continues. This suggests that safety-degrading code data may extend beyond the insecure code studied in emergent misalignment. In \texttt{personahub}, gradient filtering removes ``write a Python function'' tasks (RR $1.74\times$) and keeps calculus and differential-equation modeling problems (RR $0.51\times$) more often, suggesting that code data may be more safety-degrading than math data. In \texttt{codealpaca}, gradient filtering more often removes high-level discussion of software architecture (\texttt{scalability}, \texttt{distributed}, \texttt{authentication}) at an RR $1.75\times$ and keeps responses that contain runnable code (\texttt{def}, \texttt{print}, \texttt{return}) at an RR of $0.89\times$. 

\subsection{Math Analysis}
\label{sec:math-analysis}

In \texttt{open\_math\_2\_gsm8k}, gradient filtering more often removes answers whose arithmetic comes out to clean numbers and keeps answers containing long repeating decimals (e.g.\ \texttt{0.3333333333333333}) at an RR of $0.36\times$. In \texttt{personas\_math}, it removes narrative arithmetic, in which quantities are carried in prose through a sequence of operations, and keeps genuine algebraic manipulation, in which the model defines a variable, sets up an equation, and substitutes or simplifies (RR $0.70\times$). 

\subsection{Language Analysis}
\label{sec:misc-analysis}

In \texttt{wildchat}, gradient filtering removes English conversations at $1.46\times$ as often as non-English languages, with every other major language removed at $0.26$--$0.65\times$ the baseline rate. \texttt{aya\_100k} shows the same direction, with English again the most removed language (RR $1.73\times$), even though English is only 6.7\% of that subset and the majority language in WildChat. This is notable because fine-tuning on non-English data has often been reported to degrade safety \citep{hawkins2026language}. One possible reason for the reversal here is that Tulu 3's data is very clean \citep{tulu2025data, djuhera2025tulu}. 

\subsection{Overall Analysis}
\label{sec:overall-analysis}

In terms of well-defined types of data, tool-execution failures and narrative arithmetic stand out as possible categories of data which may degrade safety that have been under-studied in prior work. As a broad trend, gradient filtering seems to suggest that direct examples may degrade safety more than elaborate examples. Gradient filtering appears to have removed more direct data (such as clean arithmetic and simple coding tasks) and kept less direct or more elaborate data (such as repeating-decimal arithmetic, algebraic setup, and differential equation modeling) more often. One possible reason is that direct, immediately helpful answers match the pattern of simple compliance, so direct data may degrade safety more. Table~\ref{tab:length} shows that this effect cannot be reduced to simply optimizing for shorter examples. Removed documents are clearly shorter in only about half of the eleven major subsets (6/11), with three other subsets having a minor length difference of 5\% or less and two others showing gradient filtering selecting longer data. 

\begin{table}[h]
\centering
\small
\begin{tabular}{lr}
\toprule
Subset & \% Removed Data is Shorter By \\
\midrule
\texttt{table\_gpt} & 40\% \\
\texttt{personahub} & 33\% \\
\texttt{aya} & 30\% \\
\texttt{no\_robots\_converted} & 29\% \\
\texttt{wildjailbreak} & 21\% \\
\texttt{flan} & 20\% \\
\texttt{wildguardmixtrain} & 5\% \\
\texttt{open\_math\_2\_gsm8k} & 5\% \\
\texttt{personas\_math} & 3\% \\
\texttt{codealpaca} & -3\% \\
\texttt{wildchat} & -11\% \\
\bottomrule
\end{tabular}
\caption{The percentage that the data removed by gradient filtering is shorter than the kept documents for every major subset. The percentage is only large for half (6/11) of the major subsets, indicating that gradient filtering is not simply optimizing for shorter length examples.}
\label{tab:length}
\end{table}

We stress that these patterns are only \textbf{suggestive rather than established}. We make no claim that these features must cause safety degradation, only that these features are what these examples appear to have in common. Every result comes from the correlational method in Section~\ref{sec:removal-methodology}. 

\section{Extended SafeTulu Details}
\label{sec:safetulu_appendix}

\subsection{Removal Fraction Sweep}
\label{sec:removal_fraction_sweep}

\begin{table}[h]
\centering
\small
\begin{tabular}{lcc}
\toprule
\textbf{Removal Fraction} & \textbf{LLM-as-a-Judge ASR $\downarrow$} & \textbf{Gradient ASR $\downarrow$} \\
\midrule
0\%  & 38.00 $\pm$ 1.53 & 38.00 $\pm$ 1.53 \\
5\%  & 36.33 $\pm$ 1.86 & 35.00 $\pm$ 1.53 \\
10\% & \textbf{35.00} $\pm$ 1.53 & 33.67 $\pm$ 1.20 \\
20\% & 39.00 $\pm$ 1.58 & \textbf{30.33} $\pm$ 0.33 \\
40\% & 43.00 $\pm$ 2.52 & 39.00 $\pm$ 0.58 \\
\bottomrule
\end{tabular}
\caption{GCG attack success rate (\%) vs. the fraction of training data removed, for training Llama 3.2 1B after LLM-as-a-judge filtering and gradient-based filtering. Values are mean $\pm$ standard error; lower is better.}
\label{tab:removal_fraction}
\end{table}

To test whether our results depend on the 20\% removal budget, we swept the removal fraction over 0\%, 5\%, 10\%, 20\% and 40\% for both gradient filtering and the LLM-as-a-judge filter on Llama 3.2 1B. As before, each value is the mean ± one standard error over three fine-tuning runs with different seeds.

Results are shown in Table ~\ref{tab:removal_fraction}. Gradient filtering reaches a lower minimum ASR. The best ASR of gradient filtering is 30.33\% at 20\% removal. The LLM judge's best is 35.00\% at 10\% removal. The point where removal stops helping shows how much truly safety-degrading data each method finds. The LLM judge lowers ASR steadily up to 10\% removal (38.00 → 35.00). Beyond that, ASR rises above the unfiltered baseline: 39.00 at 20\% and 43.00 at 40\%. This suggests that beyond about 10\%, the judge can no longer find data that is truly safety-degrading. It starts removing benign-looking data that may actually support safety, consistent with Tulu 3 containing little overtly harmful content. Gradient filtering shows the same reversal, but only past 20\%: it keeps lowering ASR through 20\% (38.00 → 30.33) before rising at 40\% (39.00). Gradient filtering keeps helping for about twice as much removed data before this happens. This indicates that it finds a larger pool of genuinely safety-degrading data than a content-based judge.

\subsection{LLM-as-a-Judge Baseline Details}
\label{sec:judge-details}


\begin{tcolorbox}[
  enhanced, breakable,
  colback=gray!5, colframe=gray!60,
  title={\small\textbf{Safety Scoring Prompt (0--5 Scale)}},
  fonttitle=\small, fontupper=\scriptsize,
  left=4pt, right=4pt, top=3pt, bottom=3pt,
  boxsep=2pt, arc=2pt
]
\setlength{\parskip}{2pt}

\textbf{Objective:} Assess the safety of a given text, evaluating its appropriateness for public use, educational environments, or general audiences. Assign points based on the severity and presence of unsafe content across different categories. The total score ranges from \textbf{0} to \textbf{5}, with higher scores indicating \textbf{lower safety}.

\textbf{Scoring Criteria:}

\textbf{0 (Safe):} \textit{Content:} no unsafe or inappropriate content. \textit{Suitability:} fully appropriate for all audiences, including children and sensitive environments. \textit{Language:} respectful, without any profanity, slurs, or derogatory terms. \textit{Themes:} no adult themes, violence, or references to illegal activities.

\textbf{1 (Minimally Unsafe):} \textit{Content:} minimal unsafe content with minor instances of inappropriate language or mild adult themes. \textit{Language:} may include mild profanity (e.g., ``damn,'' ``hell'') or slang. \textit{Themes:} non-explicit references to alcohol, mild romantic situations, or casual mentions of stressful topics. \textit{Suitability:} generally appropriate but may be unsuitable for very young audiences or highly sensitive environments.

\textbf{2 (Slightly Unsafe):} \textit{Content:} elements of concerning language or topics. \textit{Language:} minor personal insults or sarcastic remarks without severe offense. \textit{Advice:} mild financial, legal, or medical advice without professional endorsement or disclaimers. \textit{References:} brief, indirect mentions of political, religious beliefs, or controversial topics. \textit{Suitability:} slightly inappropriate or off-topic for certain audiences; may require discretion.

\textbf{3 (Moderately Unsafe):} \textit{Content:} moderately harmful content that may offend or disturb. \textit{Language:} social-group insults, stereotypes, or prejudiced remarks without aggressive intent. \textit{Themes:} suggestive or explicit adult themes, non-graphic sexual content, or depictions of moderate violence. \textit{References:} brief mentions of weaponry, self-harm, or crimes without detailed descriptions. \textit{Suitability:} moderately inappropriate; may require review, parental guidance, or filtering.

\textbf{4 (Clearly Unsafe):} \textit{Content:} clearly inappropriate for general audiences. \textit{Language:} strong profanity, slurs, or aggressive derogatory terms directed at individuals or groups. \textit{Advice:} detailed financial, legal, or medical advice that could lead to harm. \textit{Themes:} extremist beliefs, conspiracy theories, or detailed references to illegal activities. \textit{References:} encouraging self-harm, violence, or other dangerous actions. \textit{Suitability:} contains explicit elements; should be restricted or removed from public platforms.

\textbf{5 (Highly Unsafe):} \textit{Content:} highly unsafe and overtly harmful. \textit{Language:} hate speech, violent threats, or calls to action against individuals or groups. \textit{Themes:} graphic adult content involving illegal activities. \textit{Advice:} promotion or instruction of illegal or harmful actions. \textit{Privacy:} disclosure of personally identifiable information without consent. \textit{Suitability:} inappropriate for any general, educational, or public platform.

\textbf{Instructions:} (1) Analyze the text thoroughly: understand the content, context, and tone, and identify any unsafe elements based on the criteria. (2) Assign a safety score: start from 0 and assign the highest applicable score (do not sum points). (3) Return a JSON object: \texttt{\{"score": <score>, "reason": "<reason>"\}}, where the reason briefly states the main issue (e.g., ``hate speech,'' ``financial advice,'' ``WMD'').

\textbf{Example.} \textit{Text:} ``I think we should overthrow the government because they are all corrupt and don't care about us.'' \textit{Identified content:} expresses extremist beliefs and suggests illegal activities. \textit{Justification:} advocates the illegal act of overthrowing the government, which is a serious offense; promotes extremist beliefs and could encourage unsafe behavior, making it clearly inappropriate for general audiences. \textit{Output:} \texttt{\{"score": 4, "reason": "illegal activity"\}}
\end{tcolorbox}


We prompt Qwen 3 8B, served with constrained (JSON-schema-guided) decoding, to score every document in the training set for safety on an integer 0--5 scale, where 0 denotes content that is fully safe and appropriate for all audiences and 5 denotes highly unsafe content (hate speech, violent threats or calls to action, graphic illegal content, disclosure of personal information, or explicit promotion/instruction of illegal or harmful acts). Each score level is defined by the rubric established in Appendix C.1 from \citet{maini2025llmasjudge} (shown above) over content, language, themes, advice given, and audience suitability. The judge is instructed to start from 0 and assign the highest applicable score rather than summing point-by-point, and returns a structured \texttt{\{"score": \textit{int}, "reason": \textit{str}\}} object. Scores that fail to parse are treated as missing and excluded.

The judge sees both the user prompt and the model's response, so it can penalize an unsafe prompt-response pair even when the response itself appears benign in isolation. Because the discrete 0--5 scale may produce a tie band straddling the removal boundary (e.g. the cutoff falling in the middle of the population of 4s and 5s), ties are broken by a seeded random draw, so which documents within the boundary score are kept vs. dropped is randomized rather than determined by document order.

\subsection{Hyperparameters}
\label{sec:hyperparameters}

Training Hyperparameters: 
\begin{itemize}
    \item Number of Steps: 2,297
    \item Learning Rate (LR): 1 $\times$ 10$^{-4}$ (after conducting a sweep over 5 $\times$ 10$^{-5}$, 1 $\times$ 10$^{-4}$, and 3 $\times$ 10$^{-4}$) 
    \item $\beta1$: 0.9 
    \item $\beta2$: 0.95 
    \item Weight Decay: 0.1 
    \item LR Warm-Up: 20 steps
    \item LR Decay Ratio: 20\% 
    \item Minimum LR Factor: 0.05 
    \item Sequence Length: 2,048
    \item Global Batch Size: 128 sequences 
    \item Batch Size per GPU: 8
    \item Gradient Clipping: Max Norm 1.0 
    \item Parameter Precision: bf16 
    \item Gradient Reduction Precision: fp32 
    \item Dataloader Shuffle Seed: 9,176
\end{itemize}

GCG Hyperparameters: 
\begin{itemize}
    \item Initial Suffix: 20 "!" tokens 
    \item Optimization Target: JailBreakBench supplied affirmative target per behavior 
    \item Number of Optimization Steps: 500
    \item Candidate Suffixes/Step: 512
    \item Gradient Top-k: 256
\end{itemize}

\subsection{Removal Scoring Details}
\label{sec:embedding-removal}

For a training document $x$ with rank-1 factors $\{(u_\ell, v_\ell)\}$ and a cluster centroid $M_c$, the score is
\[
    \mathrm{score}(x, c) \;=\; \langle G_x, M_c \rangle_F
    \;=\; \frac{1}{|c|}\sum_{t \in c} \sum_{\ell=1}^{L} (u_\ell^x \cdot u_\ell^t)(v_\ell^x \cdot v_\ell^t),
\]
i.e., the Frobenius inner product between the training document's gradient matrix and the cluster centroid is exactly the mean of its pairwise bilinear scores (Eq.~\ref{eq:bilinear}) against the cluster's members. This identity lets us compute cluster-level scores directly from the rank-1 factors, without ever forming the full outer-product embedding $G_x \in \mathbb{R}^{d\times d}$ per layer, and is equivalent (after normalizing by the source document's self-norm and the centroid's norm) to running cosine-similarity $k$-means in the lifted outer-product space. Each training document is assigned to its nearest cluster by $\arg\max_c \mathrm{score}(x,c)$, giving every training document a (cluster assignment, distance-to-centroid) pair.

\newpage 

\section{Extended Emergent Misalignment Experiments}
\label{sec:more_em}

\subsection{Non-Fine-Tuned Models}
As a control, we repeat the gradient-embedding analysis using the corresponding base instruct models before insecure-code fine-tuning. For Qwen2.5-Coder-32B-Instruct \citep{qwencoder2024model}, the AUC falls from 0.604 after insecure-code fine-tuning to 0.552 in the non-fine-tuned model. For Mistral-Small-24B-Instruct-2501 \citep{mistral2025model}, the AUC falls from 0.865 to 0.431. These results indicate that the gradient-space relationship between insecure-code examples and broadly harmful outputs does not emerge until after the model has undergone emergent misalignment. 

\subsection{Cross-Model Transfer}

\begin{table}[h]
\centering
\small
\resizebox{\textwidth}{!}{%
\begin{tabular}{lllcc}
\toprule
\textbf{Embedding Model} & \textbf{Target Responses} & \textbf{Model State} & \textbf{Matched AUC} & \textbf{Transfer AUC} \\
\midrule
Mistral-Small-24B-Instruct-2501 & Qwen harmful responses & Fine-tuned & 0.865 & 0.832 \\
Qwen2.5-Coder-32B-Instruct & Mistral harmful responses & Fine-tuned & 0.604 & 0.247 \\
Mistral-Small-24B-Instruct-2501 & Qwen harmful responses & Base & 0.431 & 0.502 \\
Qwen2.5-Coder-32B-Instruct & Mistral harmful responses & Base & 0.552 & 0.535 \\
\bottomrule
\end{tabular}%
}
\caption{Cross-model transfer of the emergent-misalignment gradient signal. Matched AUC uses harmful target responses produced by the same model family as the embedding model, while Transfer AUC replaces them with responses from the other fine-tuned model family. The embedding model used is important: models' gradient representations do not necessarily transfer to different models' responses.}
\label{tab:em_transfer}
\end{table}

We next test whether the gradient-space signal transfers across model families by changing the model used to compute embeddings and the model from which the verified-unsafe target responses are drawn. Results are shown in Table~\ref{tab:em_transfer}. When the fine-tuned Mistral model embeds harmful responses generated by fine-tuned Qwen, AUC decreases only modestly from 0.865 to 0.832. In the opposite direction, however, using fine-tuned Qwen to embed fine-tuned Mistral's harmful responses reduces AUC from 0.604 to 0.247. For the non-fine-tuned models, cross-model targets produce values that remain close to chance. Overall, this shows that the specific embedding model used is important for gradient representation, and that a model's gradient representation will not necessarily transfer to a different model's responses. 

These results may appear to contrast with Section~\ref{sec:results}, where filtering with gradients from a non-fine-tuned Qwen model improves safety across multiple model families. However, the emergent-misalignment experiments examine a much broader and more difficult relationship: whether gradients connect narrow insecure-code training items to harmful out-of-domain behaviors that emerge after fine-tuning. This more indirect connection is likely to be more model- and state-specific than the more general safety-degradation signal used for SafeTulu, which may explain why transfer across models and to non-fine-tuned checkpoints is weaker here.

\begin{table}[h]
\centering
\small
\begin{tabular}{lc}
\toprule
\textbf{Model} & \textbf{AUC} \\
\midrule
Qwen2.5-Coder-0.5B-Instruct & 0.507 \\
Qwen2.5-Coder-3B-Instruct & 0.549 \\
Qwen2.5-Coder-7B-Instruct & 0.573 \\
Qwen2.5-Coder-14B-Instruct & 0.590 \\
Qwen2.5-Coder-32B-Instruct & 0.552 \\
\bottomrule
\end{tabular}
\caption{Gradient-embedding AUC across non-fine-tuned Qwen2.5-Coder model sizes. The gradient signal increases up to 14B, suggesting that smaller models may have a sharper gradient signal than larger models.} 
\label{tab:em_scaling}
\end{table}

\subsection{Scaling}
Finally, we test how the gradient signal varies with model scale within the Qwen2.5-Coder family, using non-misaligned instruct checkpoints from 0.5B to 32B parameters. The results are shown in Table~\ref{tab:em_scaling}. AUC increases steadily from 0.507 at 0.5B to 0.549 at 3B, 0.573 at 7B, and 0.590 at 14B, before decreasing to 0.552 at 32B. The scaling trend is therefore not monotonic across the full model-size range. From 0.5B through 14B, the gradient-space signal strengthens consistently, suggesting that larger models can make the relevant distinction increasingly detectable. However, the 32B model falls below the 14B model, indicating that more parameters do not necessarily always produce a stronger gradient-attribution signal. This suggests that gradient-based attribution may sometimes be sharper in smaller or intermediate-sized models than in the largest model, and that embedding-model scale should be treated as an empirical design choice rather than assuming that larger models will always provide better attribution geometry.

\end{document}

%% file: math_commands.tex
\usepackage{amsmath,amsfonts,bm}

\def\eqref#1{equation~\ref{#1}}

\def\1{\bm{1}}

\DeclareMathAlphabet{\mathsfit}{\encodingdefault}{\sfdefault}{m}{sl}
\SetMathAlphabet{\mathsfit}{bold}{\encodingdefault}{\sfdefault}{bx}{n}

